\documentclass{article}
\usepackage{spconf}
\usepackage{graphicx}
\usepackage{boldline,multirow}
\usepackage{lipsum} 
\usepackage{xcolor}
\usepackage{amsfonts}
\usepackage[utf8]{inputenc}
\usepackage[T1]{fontenc}
\usepackage{amsmath} 
\usepackage{algorithm}
\usepackage{algpseudocode}
\usepackage{wrapfig}
\SetLipsumParListSurrounders{\colorlet{oldcolor}{.}\color{gray}}{\color{oldcolor}}
\usepackage{ctable} % for \specialrule command
\begin{document}
%
% \title{Multi-Domain Learning by Meta-Learning: Taking Optimal Steps in Multi-Domain Loss Landscapes by Inner-Loop Maximum a Posteriori Estimation}
\title{Multi-Domain Learning by Meta-Learning: Taking Optimal Steps in Multi-Domain Loss Landscapes by Inner-Loop Learning}
%
% \titlerunning{Multi-Domain Learning by Meta-Learning}

\name{\parbox{\linewidth}{\centering Anthony Sicilia$^1$  \quad  Xingchen Zhao$^2$ \quad Davneet S. Minhas$^3$ \quad Erin E. O’Connor$^5$\\ \textit{Howard J. Aizenstein}$^4$ \quad \textit{William E. Klunk}$^4$ \quad \textit{Dana L. Tudorascu}$^4$ \quad \textit{Seong Jae Hwang}$^{1,2}$}}
\address{
    $^{1}$Intelligent Systems Program - University of Pittsburgh \\
    Department of $^{2}$Computer Science, $^{3}$Radiology, $^{4}$Psychiatry - University of Pittsburgh \\
    $^{5}$Department of Diagnostic Radiology \& Nuclear Medicine -  University of Maryland, Baltimore}
    
\maketitle              % typeset the header of the contribution
%
%
%
%

%%%%%%%%%%%%%%%%%%%%%%%%%%%%%

\vspace{-3pt}
\begin{abstract} % up to 250 words
We consider a model-agnostic solution to the problem of Multi-Domain Learning (MDL) for multi-modal applications. Many existing MDL techniques are model-dependent solutions which explicitly require nontrivial architectural changes to construct domain-specific modules. Thus, properly applying these MDL techniques for new problems with well-established models, e.g. U-Net for semantic segmentation, may demand various low-level implementation efforts. In this paper, given emerging multi-modal data (e.g., various structural neuroimaging modalities), we aim to enable MDL purely algorithmically so that widely used neural networks can trivially achieve MDL in a model-independent manner. To this end, we consider a weighted loss function and extend it to an effective procedure by employing techniques from the recently active area of learning-to-learn (meta-learning). Specifically, we take inner-loop gradient steps to dynamically estimate posterior distributions over the hyperparameters of our loss function. Thus, our method is \textit{model-agnostic}, requiring no additional model parameters and no network architecture changes; instead, only a few efficient algorithmic modifications are needed to improve performance in MDL. We demonstrate our solution to a fitting problem in medical imaging, specifically, in the automatic segmentation of white matter hyperintensity (WMH). We look at two neuroimaging modalities (T1-MR and FLAIR) with complementary information fitting for our problem.
\end{abstract}      % 0.5 page
\vspace{-3pt}
\section{Introduction}
{\let\thefootnote\relax\footnotetext{\hspace{-10pt}Accepted to IEEE International Symposium on Biomedical Imaging 2021}}
In this paper, we consider the problem of Multi-Domain Learning (MDL) in which the goal is to take labeled data from some collection of domains $\{\mathcal{D}_{i}\}_{i}$ and minimize the risk on \textit{all} of these domains. Note, this is in contrast to the related field of Domain Adaptation (DA) which minimizes risk on only a subset of these domains referred to as the target. Although our focus is MDL, it is not uncommon for Multi-Task Learning (MTL) solutions to be applicable to MDL problems. Where MDL assumes a collection of domains $\{\mathcal{D}_{i}\}_{i}$ all paired with the same task $\mathcal{T}$, MTL assumes a collection of tasks $\{\mathcal{T}_i\}_i$ paired with a single domain $\mathcal{D}$~\cite{yang2014unified}. One simple 
%We consider the problem of Multi-Domain Learning (MDL). To understand precisely the scope of this problem, let us differentiate it from the related problems of Domain Adaptation (DA) and Multi-Task Learning (MTL). While DA takes labeled data from some collection of domains $\{\mathcal{D}_{i}\}_{i}$ and minimizes the risk on a subset of these referred to as the target, MDL, which we focus on in this paper, is slightly more general, attempting to simultaneously minimizing the risk on \textit{all} of these domains. As such, standard DA techniques such as fine-tuning may impair performance on source domains, when transferring to target domains \cite{li2017learning}. In DA, a domain is (usually) paired with a task $\mathcal{T}$, or multiple tasks $\{\mathcal{T}_i\}_i$, bringing us to compare MDL and MTL. MTL assumes a collection of tasks $\{\mathcal{T}_i\}_i$ all paired with a single domain $\mathcal{D}$, while MDL assumes a collection of domains $\{\mathcal{D}_{i}\}_{i}$ all paired with the same task $\mathcal{T}$ \cite{yang2014unified}. Although our focus is MDL, it is not uncommon for MTL solutions to be applicable to MDL problems. For example, a simple, 
model-agnostic solution to both problems comes in the form of a weighted loss function used to learn new tasks without ``forgetting'' old tasks \cite{li2017learning}. This method can be simplified and adapted for the task of MDL by specifying loss functions for each domain and jointly training on all domains by optimizing for the convex combination (weighted average) of these loss functions. Inspired by this approach, \textbf{our main contribution} is to significantly build-upon this method by dynamically estimating the optimal weights of the convex combination throughout the training process. To achieve this, we appeal to the recently growing research area of \textit{learning-to-learn} (or meta-learning) which uses the idea of \textit{hypothetical} gradient steps taken during an \textit{inner-loop optimization} to extract ``meta-information'' useful to the optimization task. Our method closely follows this idea to estimate a posterior distribution over the optimal weights of our loss function at each training iteration.

\begin{figure*}[t!]
    \centering
    \includegraphics[width=0.85\textwidth]{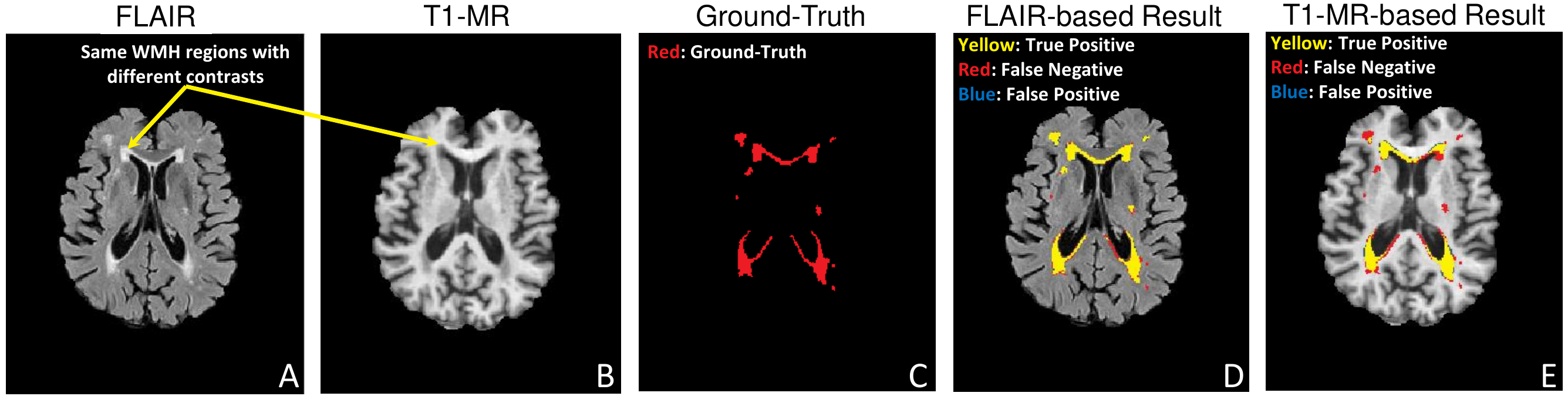}
    \caption{ \textbf{A}: FLAIR (bright WMH in the periventricular and deep white matter), \textbf{B}: Coregistered T1-MR (WMHs are less apparent), \textbf{C}: Manually segmented WMH, \textbf{D}: WMH prediction with FLAIR, E: WMH prediction T1-MR. T1-MR-based result (E), while reasonable, is still worse than FLAIR-based result (D). }
    \label{fig:image_results}
\end{figure*}
We showcase this method on a fitting problem in medical imaging, specifically, in the automatic segmentation of white matter hyperintensity (WMH) with multi-modal structural neuroimaging. Caused by various factors from neurological to vascular pathologies \cite{vermersch1996white}, WMH is prevalent in population of aging, e.g., Alzheimer's disease (AD) \cite{kandel2016white}.
% , stroke \cite{wong2002cerebral}, and bipolar disorder \cite{pillai2002increased}.
% Due to such associations, an accurate segmentation of the WMH region in structural brain images has been continuously investigated.
% Recently, as an alternative to time-consuming manual segmentation by neuroradiologists, automatic segmentation methods based on deep learning have shown promising performance \cite{rachmadi2018segmentation}.
Typically, the automatic WMH segmentation task focuses on identifying hyperintense, or bright, white matter regions in T2-weighted fluid attenuated inversion recovery (FLAIR). However, FLAIR is often acquired for neurological disorders that directly search for strokes or lesions, whereas, in observational AD studies of our focus, FLAIR is much less common and T1-weighted Magnetic Resonance (T1-MR) image is the norm. %for identifying AD-specific brain atrophies.
Unfortunately, detecting WMH in T1-MR is extremely difficult since contained WMH regions severely lack contrast -- a key feature for segmentation (see Fig.~\ref{fig:image_results} for the contrast difference and predictions). Hence, this setting provides a good opportunity for knowledge transfer across domains (T1-MR and FLAIR). While FLAIR may benefit from the higher \textit{quantity} of T1-MR samples in a given dataset, T1-MR may additionally benefit from the much higher \textit{quality} of the FLAIR samples.
Further, considering how common it is for patients to only have either T1-MR or FLAIR, MDL is particularly relevant in this case (rather than DA) to perform well on both domains (i.e., train with T1-MR \textit{and} FLAIR, but predicts well given T1-MR \textit{only}, Fig.~\ref{fig:image_results}E). In this paper, we present a solution for MDL within this context. Importantly, the approach is model-agnostic, making it easily applicable to a myriad of MDL problems besides WMH segmentation.         % 1.5 page
\vspace{-8pt}
\section{Multi-Domain Learning (MDL)}
\label{sec:rel}
Several early works on MDL are ensemble-based, combining learned domain-specific parameters into a single classifier for inference \cite{dredze2008online,dredze2010multi}. 
More recent works separate shared parameters from domain-specific parameters via residual adapters \cite{rebuffi2017learning} or a two-sided network \cite{yang2014unified}. Even adversarial approaches have been proposed \cite{schoenauer2019multi} which requires separating the model parameters into a feature extractor and a task-specific network. Notably, at some level, these methods are all \textit{model-dependent}, requiring explicit changes to the network architecture.
This is less desirable if one wishes to enable MDL in segmentation since standard existing methods may not be trivially applicable to U-Net \cite{RonnebergerFB15}
(e.g., for adversarial approaches, one still needs to somehow define where feature-extraction stops and classification begins within the U-Net structure). 
Conversely, our approach is \textit{model-agnostic}, making no model-dependent changes. It is therefore applicable to most existing models (including U-Net). This flexibility adds a great practical value for the end-user who wishes to enable MDL in a ``plug-and-play'' manner.

\vspace{0.5em}
\noindent\textbf{Learning to Learn.}
Learning-to-learn (meta-learning) is an algorithmic effort to not only learn some set of model parameters, but to learn the best \textit{way} in which those model parameters can be learned. %explicitly train the model to be optimal at certain tasks.
%In our own context, a model with meta-learned training for multiple domains should perform better than the same exact model without meta-learned training.
Many recent popularizations of this concept \cite{andrychowicz2016learning,finn2017model}
-- and formalizations \cite{grefenstette2019generalized} -- 
largely involve an \textit{inner}- and \textit{outer-loop}. The dual-loop scheme uses the \textit{inner-loop} to extract hypothetical model performance \textit{if} the model were optimized in some way. From this, in the \textit{outer-loop}, the hyperparameters of interest (e.g., the way the model is optimized) can be updated \cite{andrychowicz2016learning,finn2017model}, or the model itself can be updated in a modified way \cite{li2018learning}. 
While it is evident that meta-learning is an active area of research, our method focuses on the special case of a weighted loss function for MDL. Unlike many meta-learning solutions in the MTL problem space~\cite{finn2017model,bechtle2019meta,sung2017learning}, we have only a single task, making it unclear how we could pre-train our hyperparameters as usual (i.e., using a distribution over tasks). To combat this, we exploit the functional form of our loss-function and use Bayesian estimation techniques to dynamically estimate our hyper-parameters during training (i.e., \textit{without} any pre-training phase). One fallout of this, is an interesting differentiation of our approach from existing meta-learning literature. While the majority of meta-learning solutions are fully gradient based, our technique, instead, uses MAP estimation during inner-loop optimization.
% To combat this, we exploit the functional form of our loss-function and use Bayesian estimation techniques to dynamically estimate our hyper-parameters during training (i.e., \textit{without} any pre-training).

        % 1 page
\vspace{-10pt}
\section{Proposed Approach}
\label{sec:meth}
\begin{figure}
    \centering
    \includegraphics[width=0.65\columnwidth]{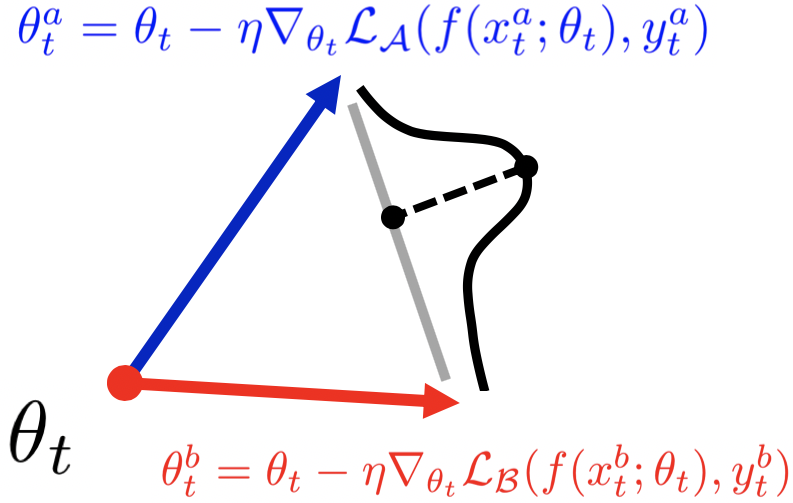}
  \caption{\label{fig:geom}\textbf{Proposed Approach.} We learn a distribution over the \textit{optimal} convex-combination of the two gradient directions, selecting the most likely to be optimal.}
\end{figure}

We describe our approach (Alg.~\ref{alg:one}, Fig.~\ref{fig:geom}) which can be applied universally to nearly any neural network model without model-specific changes.
Our meta-learning procedure with outer- and inner-loop is as follows: (i) \textbf{outer-loop} updates the model parameters $\theta$ based on (ii) \textbf{inner-loop} which learns and updates our hyperparameter ($\lambda_\theta$).
We first formalize the weighted loss function used in the outer-loop.
Ultimately, we interpret this loss as an expectation over an \textit{optimal update choice}, allowing us to learn $\lambda_\theta$ by MAP estimation.

\subsection{Outer-loop Optimization of Model Parameter $\theta$}
We define a \textit{domain} $\mathcal{D} = (\mathcal{X}, p(x))$ as a feature space $\mathcal{X}$ paired with a distribution of samples from that space $p(x)$~\cite{cheplygina2019not}. For the remainder of the paper, we generally assume only two domains $\mathcal{A} = (\mathcal{X_A}, p_\mathcal{A}(x))$ and $\mathcal{B}= (\mathcal{X_B}, p_\mathcal{B}(x))$. We do this for brevity and for our two domain neuroimaging application, but in a subsequent section, we indeed show an easy extension to more than two domains. Further, we assume a single task $\mathcal{T} = (\mathcal{Y}, q(y))$ (e.g., segmentation), a pre-specified model $f$ (e.g., U-Net), and a possibly domain-specific 
% functions of risk $\mathcal{R_A}$ and $\mathcal{R_B}$; i.e., 
% \begin{equation}
% \mathcal{R}_\cdot(\theta) = \mathbb{E}_{x_i\sim p_\cdot(x), y_i\sim q(y)}\mathcal{L_A}(f(x_i; \theta), y_i)\end{equation}
loss function for both $\mathcal{A}$ and $\mathcal{B}$ written $\mathcal{L_A}$ and $\mathcal{L_B}$ respectively. The goal of our method is to dynamically determine the optimal weighting of these losses. Specifically, we seek $\lambda_\theta$ with $0 \leq \lambda_\theta \leq 1$ for the training objective below
\begin{equation} 
\label{eqn:one}
\lambda_\theta \mathcal{L_A}(f(x^a; \theta), y^a) + (1 - \lambda_\theta ) \mathcal{L_B}(f(x^b; \theta), y^b) 
\end{equation}
where $\theta$ is the current model parameters. The mini-batches $(x^a, y^a)\sim (p_\mathcal{A}(x), q(y))$ and $(x^b, y^b)\sim (p_\mathcal{B}(x), q(y))$ are (input,label) pairs from domains $\mathcal{A}$ and $\mathcal{B}$ respectively.
% the mini-batches are drawn as follows $(x^a, y^a) \sim (p_\mathcal{A}(x), q(y))$, $(x^b, y^b) \sim (p_\mathcal{B}(x), q(y))$.
In practice, this objective is achieved using a modified SGD to update $\theta$. In particular, at step $t$, we set $\theta_{t+1}$ as below
\begin{equation}
\label{eqn:update}
\begin{split}
\theta_{t+1} \gets \theta_t - \eta \nabla_{\theta_t} & \left [\lambda_{t} \mathcal{L_A}(f(x_t^a; \theta_t), y_t^a) \right .
\\ & \left . + (1 - \lambda_{t} ) \mathcal{L_B}(f(x_t^b; \theta_t), y_t^b) \right ]
\end{split}
\end{equation}
with $\eta$ the learning rate and mini-batches $(x_t^a, y_t^a)$ and $(x_t^b, y_t^b)$.
Since Eq.~\eqref{eqn:update} involves two losses, the learned $\lambda_{t}$ weights the effect of gradients $\nabla_{\theta_t}\mathcal{L_A}$ and $\nabla_{\theta_t}\mathcal{L_B}$ which are best for $\mathcal{A}$ and $\mathcal{B}$ respectively.
% Since Eq.~\eqref{eqn:update} involves two losses ($\mathcal{L_A}$ and $\mathcal{L_B}$), the learned $\lambda_{t}$ weights the influence of the gradient direction best for domain $\mathcal{A}$, $\nabla_{\theta_t}\mathcal{L_A}(f(x_t^a; \theta_t), y_t^a)$, and best for domain $\mathcal{B}$, $\nabla_{\theta_t}\mathcal{L_B}(f(x_t^b; \theta_t), y_t^b)$.
This \textbf{outer-loop} optimization (Alg.~\ref{alg:one} line 10) differs from standard SGD with weighted losses since $\lambda_t$ depends on the \textit{current} $\theta_t$ rather than being fixed. In the next section, we describe the inner-loop optimization step which estimates $\lambda_t$ dynamically.

\algrenewcommand\algorithmicindent{0.5em}%
\begin{algorithm}[t!]\small 
    \caption{Approach using the \textbf{conservative} configuration}
    \label{alg:one}
    \hspace*{\algorithmicindent} \textbf{Domain $\mathcal{A}$ input, labels, loss: $x^a, y^a, \mathcal{L_A}$} \\
    \hspace*{\algorithmicindent} \textbf{Domain $\mathcal{B}$ input, labels, loss: $x^b, y^b, \mathcal{L_B}$} \\
    \hspace*{\algorithmicindent} \textbf{Model Parameters, Loss Weighting, Learning-Rate:} $\theta, \lambda, \eta$
    % \hspace*{\algorithmicindent} \textbf{:} $\lambda_{\theta}$ \\
    % \hspace*{\algorithmicindent} \textbf{:} $\eta$ 
    \begin{algorithmic}[1] % The number tells where the line numbering should start
        \Procedure{MetaLearningForMDL}{}
            \For{mini-batch $t$}
            \State{Split $x^a_t, y^a_t, x^b_t, y^b_t$ into $\dot{x}^a_t, \dot{y}^a_t, \dot{x}^b_t, \dot{y}^b_t$ and $\ddot{x}^a_t, \ddot{y}^a_t, \ddot{x}^b_t, \ddot{y}^b_t$}
            \State{$\theta^a_t \gets \theta_t - \eta\nabla_{\theta_t}\mathcal{L_A}(f(\dot{x}_t^a; \theta_t), \dot{y}_t^a)$ \Comment{Inner-Loop for $\mathcal{A}$}}
                \State{$\theta^b_t \gets \theta_t - \eta\nabla_{\theta_t}\mathcal{L_B}(f(\dot{x}_t^b; \theta_t), \dot{y}_t^b)$\Comment{Inner-Loop for $\mathcal{B}$}}
                \State{$H^t_\mathcal{A} \gets \mathcal{L_A}(f(\ddot{x}_t^a; \theta^a_t), \ddot{y}_t^a) + \mathcal{L_B}(f(\ddot{x}_t^b; \theta^a_t), \ddot{y}_t^b)$}
                \State{$H^t_\mathcal{B} \gets \mathcal{L_A}(f(\ddot{x}_t^a; \theta^b_t), \ddot{y}_t^a) + \mathcal{L_B}(f(\ddot{x}_t^b; \theta^b_t), \ddot{y}_t^b)$}
                %\State{\hspace{10pt}\textbf{if} $H^t_\mathcal{A} < H^t_\mathcal{B}$ \textbf{then} $\Lambda_t \gets 1$ \textbf{else} $\Lambda_t \gets 0$}
                % \If{$H^t_\mathcal{A} < H^t_\mathcal{B}$ }
                % \State{$\Lambda_t \gets 1$}
                % \Else
                % \State{$\Lambda_t \gets 0$}
                % \EndIf
                \State{$N_t \gets \sum_{i=t-T}^{t} \Lambda_{i}(H^i_\mathcal{A} < H^i_\mathcal{B})$}
                \State{$\lambda_{t} \gets \frac{\alpha + N_t - 1}{\alpha + \beta + T - 2}$ \Comment{MAP Estimate for  $\lambda_t$}}
                \State{$\theta_{t+1} \leftarrow$ Update via Eq.~\eqref{eqn:update}}
            % \State{$\theta_{t+1} \gets \theta_t- \eta \nabla_{\theta_t} \left [\lambda_{t} \mathcal{L_A}(f(x_t^a; \theta_t), y_t^a) + (1 - \lambda_{t} ) \mathcal{L_B}(f(x_t^b; \theta_t), y_t^b) \right ]$}
            \EndFor
        \EndProcedure
    \end{algorithmic}
\end{algorithm}

\subsection{Inner-Loop Optimization of Hyperparameter $\lambda_{\theta}$}

\noindent\textbf{MAP Estimation of $\lambda_{\theta}$.}
We now discuss how to pick the \textit{optimal} $\lambda_\theta$ at each time step.
% Before we describe the definition of \textit{best}, for now, we assume some notion of an \textit{optimal update choice} is given and that this choice boils down to taking a step in the direction best for $\mathcal{A}$ \textit{or}  $\mathcal{B}$.
Here, the definition of optimal is fairly involved and formally analyzing how we define optimal requires a more complete picture. Subsequently, we assume for now that some notion of an \textit{optimal update choice} is given and that this choice boils down to taking a step in the direction best for domain $\mathcal{A}$ (e.g., by using $\nabla_{\theta_t}\mathcal{L_A}$) or best for domain $\mathcal{B}$ (e.g., by using $\nabla_{\theta_t}\mathcal{L_B}$).
Under this assumption, it is straightforward to interpret the multi-domain loss in Eq.~\eqref{eqn:one} as an expectation over the \textit{optimal update choice} (i.e., the expected \textit{best} gradient to choose at time $t$). To see this, we assume during the update process there exists a sequence of (not necessarily i.i.d.) Bernoulli random variables indicating whether a step in the direction best for domain $\mathcal{A}$ or $\mathcal{B}$ is optimal. We can write the sequence $(\Lambda_t)_t$ where $t$ indexes over the sequential update process given in Eq.~\eqref{eqn:update}, $\Lambda_{t} \sim \mathrm{Bernoulli}(\lambda_{t})$, and $\Lambda_{t} = 1$ represents the event that taking a gradient step in the direction $\nabla_{\theta_t}\mathcal{L_A}$ is optimal.

It then becomes simple to optimize $\lambda_t$ dynamically by assuming a prior and updating sequentially with Maximum a Posteriori (MAP) Estimation. To meet the requirements of MAP Estimation, we make the simplifying assumption that the $\Lambda_t$ are i.i.d. in a small temporal window of size $T$ (i.e., we perform our MAP updates using a history of length $ \leq T$). Thus, we can, as usual,\footnote{\textit{usual}, because a Beta Distribution is the Bernoulli Distribution's conjugate prior} assume the Beta($\alpha, \beta$) as our prior over $\lambda_t$ and explicitly compute the MAP estimate (Alg.~\ref{alg:one} line 8-9). Since the MAP estimate is precisely the mode of the posterior distribution, taking the log-likelihood and differentiating gives
\begin{equation}
\lambda_t = \frac{\alpha + N_t - 1}{\alpha + \beta + T - 2}
% \lambda_t = ({\alpha + N_t - 1})/({\alpha + \beta + T - 2})
\end{equation}
where $T$ is the history length and $N_t = \sum_{i=t-T}^{t} \Lambda_t$.

\vspace{0.5em}
\noindent\textbf{Defining the Optimal Update Choice.}
As alluded to, we still need define the \textit{optimal update choice}. In particular, this implies we must define when $\Lambda_t = 1$, or equivalently, when it is best to take a step in the direction $\nabla_{\theta_t}\mathcal{L_A}$. With no prior knowledge, it is not clear when this should be the case. But, during training, we propose to effectively ``explore'' the local properties of our optimization space to gain the needed insight. This may be done using meta-learning. Specifically, during \textit{inner-loop} phase, we can compare model performance after computing hypothetical gradient steps favoring $\mathcal{A}$ and $\mathcal{B}$, respectively. In the case of domain $\mathcal{A}$, we randomly split the mini-batch $x_t^a, y_t^a$ into a \textit{meta-train} set $\dot{x}_t^a,\dot{y}_t^a$ and \textit{meta-test} set $\ddot{x}_t^a,\ddot{y}_t^a$ (Alg.~\ref{alg:one} line 3). Then, we compute the hypothetical gradient favoring $\mathcal{A}$ (Alg.~\ref{alg:one} line 4)
\begin{equation} 
\theta^a_t = \theta_t - \eta\nabla_{\theta_t}\mathcal{L_A}(f(\dot{x}_t^a; \theta_t), \dot{y}_t^a),
\end{equation}
and the hypothetical loss favoring $\mathcal{A}$ (Alg.~\ref{alg:one} line 6)
\begin{equation} 
\label{eqn:hyp}
H^t_\mathcal{A} = \mathcal{L_A}(f(\ddot{x}_t^a; \theta^a_t), \ddot{y}_t^a) + \mathcal{L_B}(f(\ddot{x}_t^b; \theta^a_t), \ddot{y}_t^b).
\end{equation}
We can similarly arrive at $\theta^b_t$ and $H^t_\mathcal{B}$ (Alg.~\ref{alg:one} line 5 and 7). 

Given this information, we can more formally define two optimal update choices: 
\begin{enumerate}
    \item \textbf{greedy}: $\Lambda_t = 1$ if $H^t_\mathcal{B} > H^t_\mathcal{A}$, otherwise $\Lambda_t = 0$
    \item \textbf{conservative}: $\Lambda_t = 1$ if $H^t_\mathcal{A} >H^t_\mathcal{B}$, otherwise $\Lambda_t = 0$
\end{enumerate}
Note, the conservative definition is used in Alg.~\ref{alg:one} line~8. To gain insight on these definitions, we must do some analysis. In particular, these hypothetical losses are functions of the model parameters, and so, we can analyze them by looking at the dominant terms in their Taylor Expansions (centered at $\theta_t$). Note, this type of analysis is common for interpretation of the inner-loop \cite{reptile,li2018learning}. 
% In the case of $H^t_\mathcal{A}$, when we evaluate at $\theta_t$, we have the following to be true within a small neighborhood
So, in applying the analysis to $H^t_\mathcal{A}$ evaluated at $\theta_t^a$, the following is true\footnote{Since meta-train/test sets are simply samples drawn from the same distribution, we de-identify them in this expansion for interpretation.} for small enough $\eta$
\begin{equation} 
\begin{aligned}
H^t_\mathcal{A} \ & = \ \mathcal{L_A}(f(x_t^a; \theta_t), y_t^a) + \mathcal{L_B}(f(x_t^b; \theta_t, y_t^b) \\ & - \eta\nabla_{\theta_t}\mathcal{L_A}(f(x_t^a; \theta_t), y_t^a)^\mathrm{T}\nabla_{\theta_t}\mathcal{L_A}(f(x_t^a; \theta_t), y_t^a) \\&-\eta\nabla_{\theta_t}\mathcal{L_B}(f(x_t^b; \theta_t), y_t^b)^\mathrm{T}\nabla_{\theta_t}\mathcal{L_A}(f(x_t^a; \theta_t), y_t^a)\\
& + O(\eta^2).
\end{aligned}
\end{equation}
Now, if we apply the same analysis to $H^t_\mathcal{B}$, we notice there are multiple common terms in the Taylor Expansions. So, if we ignore $O(\eta^2)$ terms (which are small) and recognize the definition of the L2 norm, we have an approximation of $H^t_\mathcal{A} - H^t_\mathcal{B}$ as below
\begin{equation}
\begin{split}
H^t_\mathcal{A} - H^t_\mathcal{B} \ & \approx \ \eta ||\nabla_{\theta_t}\mathcal{L_B}(f(x_t^b; \theta_t), y_t^b)||_2^2 \\ & -  \eta||\nabla_{\theta_t}\mathcal{L_A}(f(x_t^a; \theta_t), y_t^a)||_2^2.
\end{split}
\end{equation}
Hence, in the \textbf{greedy} definition with $H^t_\mathcal{B} > H^t_\mathcal{A}$, we can infer 
\begin{equation}
||\nabla_{\theta_t}\mathcal{L_A}(f(x_t^a; \theta_t), y_t^a)|| > ||\nabla_{\theta_t}\mathcal{L_B}(f(x_t^b; \theta_t), y_t^b)||. 
\end{equation}
Likewise, in the \textbf{conservative} definition with $H^t_\mathcal{A} > H^t_\mathcal{B}$, we can infer \begin{equation}
    ||\nabla_{\theta_t}\mathcal{L_B}(f(x_t^b; \theta_t), y_t^b)|| > ||\nabla_{\theta_t}\mathcal{L_A}(f(x_t^a; \theta_t), y_t^a)||.
\end{equation}
Since $\lambda_t$ is the probability that $\nabla_{\theta_t}\mathcal{L_A}$ is the optimal update choice, we see that the {greedy} definition prefers larger gradient steps, while the {conservative} definition prefers smaller.

\vspace{.5em}
\noindent\textbf{More than Two Domains.}
Generalizing our approach to more than two domains is straightforward. Eq.~\eqref{eqn:one} is extended to a convex combination with additional weights for each added domain. Next, the sequence of Bernoulli Distributions becomes a sequence of Multinomial Distributions whose conjugate prior is a Dirichlet; the MAP Estimate is still analytic. Lastly, the \textit{optimal update choice} (Alg.~\ref{alg:one} line 8) is defined by $\mathrm{argmax}$ instead of $>$ and $\mathrm{argmin}$ instead of $<$.        % 2 page
\vspace{-3pt}

\section{Experiments}
\label{sec:res}
\subsection{Data and Preprocessing}
We randomly selected $N$=20 older participants with WMH from our local normal aging AD study who were cognitively normal at the time of scan with mean age of $81.2$ (s.d.$=7.15$), 14 females, and a mean education of 14.2 (s.d.$=2.44$) years.
For each subject, we used a 3T Siemens Trio TIM scanner and 12-channel head coil to collect T1-MR (TE=2.98ms, TR=2.3s, FA=9$^\circ$, 1$\times$1$\times$1.2mm voxel) and FLAIR (TE=90ms, TR=9.16s, FA=150$^\circ$, 1$\times$1$\times$3mm voxel). For each pair of T1-MR and FLAIR, we used FSL \cite{muschelli2015fslr} to process them in the following order: (a) spatially align T1-MR to FLAIR (212$\times$256$\times$48   dims), (b) N4-correction \cite{tustison2010n4itk}, (c) skull-strip using FSL BET, and (d) intensity normalize using WhiteStripe \cite{shinohara2014statistical}.
The ground-truth WMH in each FLAIR was labeled by a neuroradiologist on 5 continuous and identical slices across the subjects where WMH is common. 

\subsection{Experiment Setup}
% \noindent\textbf{Base Models.}
We use two base networks: (i) the standard \textbf{U-Net} \cite{RonnebergerFB15} and (ii) a \textit{light-weight} (\textbf{LW}) variant of U-Net with 3\% of the parameters and no pooling layers or skip-connects.
% (i.e. it down-samples the image using only learned convolutions without padding and up-samples the learned feature representation using learned transpose convolutions).
% For each of these baseline models, we apply our approach with the following settings \textit{without} any architecture modifications.

\vspace{0.5em}
\noindent\textbf{Our Methods.}
We setup our methods as described in Section \ref{sec:meth} with FLAIR for $\mathcal{A}$ and T1-MR for $\mathcal{B}$. We try $T=\{25,100\}$ for both the greedy (\textbf{Ours-G-}$T$) and conservative (\textbf{Ours-C-}$T$) versions. We use a Beta(5,5) as our prior for $\lambda_t$; this assumes equal likelihood for FLAIR/T1-MR to be optimal and imposes low likelihood of 0 or 1. Again, these are applied to the base models (U-Net, LW) \textit{without} any architecture changes in a completely model-agnostic manner.

\vspace{0.5em}
\noindent\textbf{Other Baselines.}
% As our main goal in this paper is to demonstrate a straightforward, model-agnostic method for improving performance in MDL.
% Since our method focuses on developing a technique to dynamically estimate hyperparameters of a weighted loss-function, our baselines focus on alternatives to dynamically estimating (i.e. picking a fixed weight). 
% We also include several baselines for simple but reasonable settings of hyperparameters to emphasize the importance of our MAP estimation. 
The baselines are applied to both U-Net and LW as follows: {\bf (1) F50-T50}: Fix the weighting of both FLAIR and T1-MR at 0.5 to treat them equally. This is the most na\"ive way to use any models without considering MDL. { \bf (2) F10-T90}: Fix the weighting of FLAIR at 0.10 and T1-MR at 0.90, largely favoring T1-MR.  { \bf (3) F90-T10}: Fix the weighting of FLAIR at 0.90 and T1-MR at 0.10, largely favoring FLAIR. {\bf (4) Simple}: Heuristically update the hyperparameter $\lambda_\theta$ in Eq.~\eqref{eqn:one} proportional to the difference of the hypothetical losses: $\lambda_{t+1} = \lambda_{t} + \gamma (H^t_{\mathrm{FLAIR}} - H^t_{\mathrm{T1}})/{|H^t_{\mathrm{FLAIR}}|}$.
We set {\bf Simple-G} with $\gamma = -0.1$ and {\bf Simple-C} with $\gamma = 0.1$ to heuristically mimic \textbf{Ours-G} and \textbf{Ours-C} respectively.
% \begin{enumerate}
% \vspace{-5pt}
%     \item { \bf F50-T50}: We fix the weighting of both FLAIR and T1-MR at 0.5 to treat them equally, which would be the most na\"ive way to use any models without the consideration of MDL.
%     % providing a baseline of how weighted loss-functions perform in general.
%     % This would be the most basic way to use any models (i.e., FLAIR and T1-MR are treated equally).
%     \item { \bf F10-T90}: We fix the weighting of both FLAIR at 0.10 and T1-MR at 0.90 providing an extreme favoring of T1-MR. 
%     \item { \bf F90-T10}: We fix the weighting of both FLAIR at 0.90 and T1-MR at 0.10 providing an extreme favoring of FLAIR.
%     \item {\bf Simple}: We provide a heuristic alternative to our MAP estimation by updating the weight $\lambda_\theta$ as in Eq.~\eqref{eqn:one} by an amount proportional to the difference between the hypothetical losses: $\lambda_{t+1} = \lambda_{t} + \gamma (H^t_{\mathrm{FLAIR}} - H^t_{\mathrm{T1}})/{|H^t_{\mathrm{FLAIR}}|}$.
%     % \begin{equation} \small
%     %     \lambda_{\theta_{t+1}} = \lambda_{\theta_{t}} + \gamma (H^t_{\mathrm{FLAIR}} - H^t_{\mathrm{T1}})/{|H^t_{\mathrm{FLAIR}}|}
%     % \end{equation}
%     Thus, we set {\bf Simple-G} with $\gamma = -0.1$ and {\bf Simple-C} with $\gamma = 0.1$ to heuristically mimic \textbf{Ours-G} and mimic \textbf{Ours-C} respectively.
%     % We change the sign primarily to allow for both a greedy and conservative alternative (see Section \ref{sec:meth}) as is available in our proposed method.
% \vspace{-5pt}
% \end{enumerate}

\vspace{0.5em}
\noindent\textbf{Loss Function.} For both FLAIR and T1-MR we minimize the sum of the cross-entropy and dice score loss -- a differentiable version of the Dice Score (perhaps) first proposed in~\cite{MilletariNA16}. We use the DSC loss variant found in \cite{DBLP:journals/corr/abs-1802-10508}.

\vspace{0.5em}
\noindent\textbf{Simulating Variation in Data-Availability.}
To show the efficacy of our method when the number of training subjects with FLAIR is reduced, we explore randomly down-sampling the number of subjects who have FLAIR during training. We try all FLAIR subjects (\textbf{12F}) and 2/3 of FLAIR subjects (\textbf{8F}).% where $K$F indicates $K$ number of FLAIR subjects in the training set.
% We show results when the number of FLAIR equals the number of T1-MR (i.e. with 3-folds for training this number is 12), when the number of subjects with FLAIR during training is only 8, and finally when the number of subjects is only 4. Under this setup, at the start of training for any model, we randomly up-sample FLAIR slices from \textit{only} the available subjects to re-balance our dataset.

\vspace{0.5em}
\noindent\textbf{Training Details. }
We use SGD with an initial learning rate of $0.01$ (multiply by $0.1$ if no validation improvement for 20 epochs and stop after 50 epochs of no improvement).
% We decay this learning rate when performance on the validation fold plateaus for greater than 20 epochs and further use early-stopping, terminating training when validation performance plateaus for greater than 50 epochs. 
We randomly augment each training slice by rotation, shearing, and scaling. Each mini-batch of size 8 is randomly sampled from both FLAIR and T1-MR.
%slices without imposing even ratios. Randomness mimics domain-uncertainty at test-time  and improves results for all configurations.
% Without within-batch domain-count randomization, the means and the variances of these layers would represent a distribution that is \textit{consistently} unlike the distribution at test-time (where only a single-domain would be input). With randomization, this \textit{consistency} is removed and cannot be exploited (overfit too).
% We found this randomization to be important to the training process to avoid significant over-fitting; we hypothesize this is due to Batch Normalization layers throughout the network.
% Without within-batch domain-count randomization, the means and the variances of these layers would represent a distribution that is \textit{consistently} unlike the distribution at test-time (where only a single-domain would be input). With randomization, this \textit{consistency} is removed and cannot be exploited (overfit too).
For each setup, we use 5-fold CV (12 train, 4 validate, 4 test) and compute the mean and standard deviations over 5 repeated runs on NVIDIA RTX2080Ti. Code will be made available upon publication.
% That is, at the start of experimentation, we randomly separate our 20 subjects into 5 folds of 4. For each fold, during training, we select 3 of the remaining folds to use for training and 1 for validation.
% Numbers shown are averaged over folds when each was held-out from training in this fashion. 
% \vspace{3pt}\noindent\textbf{Code and Hardware. }
% We implement all experiments using the widely used python library PyTorch \cite{NEURIPS2019_9015} and additionally use the recently developed meta-learning extension Higher \cite{grefenstette2019generalized}. {\color{red} GPU details and training time.}
% We used PyTorch \cite{NEURIPS2019_9015} and a meta-learning extension \cite{grefenstette2019generalized} with NVIDIA RTX2080Ti.

%Models and setups
%cross-validation, train/val/test split, u-net setup/data augmentation, learning rate, batch size, meta-learning setup, loss function, train only on 5 slices with manual tracing, GPU, Pytorch, approximate training time, etc.

\begin{figure}
\centering
\includegraphics[width=.8\columnwidth]{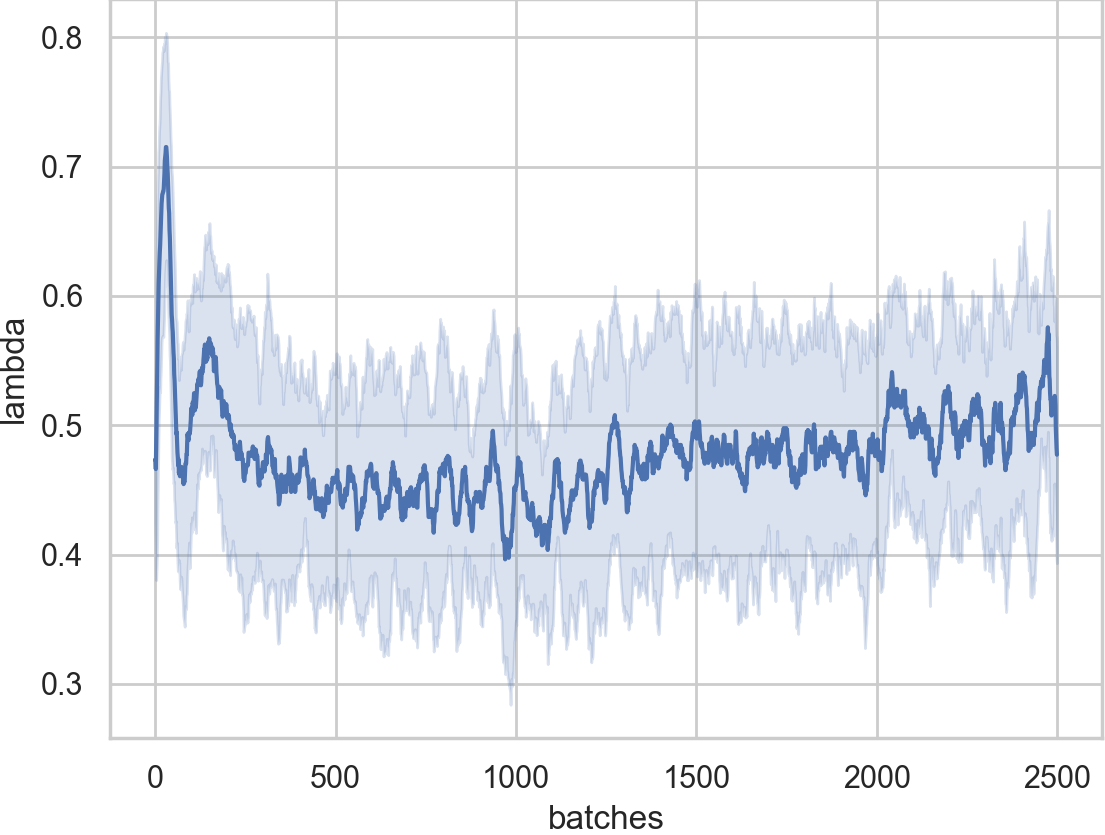}
\caption{\label{fig:lambdas}Visualization of how the optimal update choice  $\lambda_\theta$ changes over time (U-Net 8F). Line shows mean. Band shows s.d. Early spikes favor more informative FLAIR samples.
% Visualization of how Ours-C-25 modifies $\lambda_\theta$ over time (U-Net 8F). Solid line shows mean. Band shows s.d. Early spikes favor FLAIR, followed by T1-MR, then gradual return to even favor.
% , selecting the most likely to be optimal.
}
\end{figure}

\begin{table}[!t]
\caption{\label{table:full} Means and standard deviations (s.d.) of metrics across all setups using two models (U-Net and LW) and two numbers of FLAIR subjects (12 and 8). 
% For LW, we show baseline and best performing configurations with full results in Supplement.
-F and -T indicate the metrics are computed over FLAIR and T1-MR samples respectively. GAIN-$\mu$ is the total (summed) increase in DSC over the baseline F50-T50.  GAIN-$\sigma$ is the total decrease in s.d. of DSC from F50-T50. Results indicate our method increases DSC while reducing the s.d., in particular, when FLAIR data-availability is reduced.}
\centering
% \scriptsize\renewcommand{\arraystretch}{1}%\setlength{\tabcolsep}{3pt}
\resizebox{\columnwidth}{!}{%
\centering 

\setlength{\tabcolsep}{7pt}
\renewcommand{\arraystretch}{1.0}
\begin{tabular}{l|cc||cc}
\specialrule{.1em}{.1em}{0.1em} 
% Model     & DSC-F & DSC-T & PPV-F & PPV-T & AUC-F & AUC-T  \\ \hline\hline
\textbf{LW} (12F)  & DSC-F & DSC-T & GAIN-$\mu$ &  GAIN-$\sigma$\\\hline % & AUC-F & AUC-T            \\ \hline
F50-T50   & 0.757 $\pm$ 0.011 & 0.360 $\pm$ 0.031 & 0.0 & 0.0\\% & 0.983 $\pm$ 0.004 & 0.952 $\pm$ 0.007\\
F10-T90   & 0.729 $\pm$ 0.006 & 0.404 $\pm$ 0.026 & 0.016 & 0.010\\% & 0.977 $\pm$ 0.006 & 0.954 $\pm$ 0.006\\
F90-T10   & 0.766 $\pm$ 0.008 & 0.278 $\pm$ 0.033 & -0.073 & 0.001\\% & 0.986 $\pm$ 0.004 & 0.946 $\pm$ 0.007\\\hline
Simple-G   & 0.740 $\pm$ 0.023 & 0.152 $\pm$ 0.072 & -0.225 & -0.053\\% & 0.977 $\pm$ 0.011 & 0.880 $\pm$ 0.052\\
Simple-C   & 0.714 $\pm$ 0.062 & 0.325 $\pm$ 0.050 & -0.078 & -0.070\\% & 0.976 $\pm$ 0.010 & 0.941 $\pm$ 0.020\\\hline
Ours-G-25  & 0.758 $\pm$ 0.009 & 0.366 $\pm$ 0.029 & 0.007 & 0.004\\% & 0.983 $\pm$ 0.003 & 0.952 $\pm$ 0.003\\
Ours-G-100 & 0.759 $\pm$ 0.010 & 0.375 $\pm$ 0.028 & \textbf{0.017} & 0.004 \\%& 0.983 $\pm$ 0.003 & 0.952 $\pm$ 0.006\\
Ours-C-25  & 0.758 $\pm$ 0.007 & 0.356 $\pm$ 0.025 & -0.003 & 0.010\\% & 0.983 $\pm$ 0.003 & 0.953 $\pm$ 0.005\\
Ours-C-100 & 0.755 $\pm$ 0.008 & 0.351 $\pm$ 0.018 & -0.011 & \textbf{0.016}\\% & 0.982 $\pm$ 0.004 & 0.951 $\pm$ 0.008\\ \hline\hline  
\hline\textbf{LW} (8F)  & DSC-F & DSC-T & GAIN-$\mu$ &  GAIN-$\sigma$\\\hline % & AUC-F & AUC-T            \\ \hline\hline
F50-T50   & 0.753 $\pm$ 0.008 & 0.361 $\pm$ 0.023 & 0.0 & 0.0\\% & 0.982 $\pm$ 0.004 & 0.952 $\pm$ 0.007\\
F10-T90   & 0.725 $\pm$ 0.008 & 0.393 $\pm$ 0.026 & 0.004 & -0.003\\% & 0.975 $\pm$ 0.006 & 0.95 $\pm$ 0.009\\
F90-T10   & 0.766 $\pm$ 0.013 & 0.291 $\pm$ 0.030 & -0.057 & -0.012\\% & 0.986 $\pm$ 0.004 & 0.948 $\pm$ 0.007\\\hline
Simple-G   & 0.738 $\pm$ 0.020 & 0.152 $\pm$ 0.055 & -0.224 & -0.044\\% & 0.978 $\pm$ 0.009 & 0.881 $\pm$ 0.067\\
Simple-C   & 0.716 $\pm$ 0.063 & 0.311 $\pm$ 0.081 & -0.087 & -0.113\\% & 0.975 $\pm$ 0.014 & 0.935 $\pm$ 0.033\\\hline
Ours-G-25  & 0.755 $\pm$ 0.007 & 0.361 $\pm$ 0.023 & 0.002 & 0.001\\% & 0.982 $\pm$ 0.004 & 0.952 $\pm$ 0.007\\
Ours-G-100 & 0.756 $\pm$ 0.010 & 0.368 $\pm$ 0.030 & \textbf{0.010} & -0.009 \\% & 0.984 $\pm$ 0.004 & 0.956 $\pm$ 0.005\\
Ours-C-25  & 0.752 $\pm$ 0.007 & 0.355 $\pm$ 0.021 & -0.007 & \textbf{0.003}\\% & 0.983 $\pm$ 0.006 & 0.953 $\pm$ 0.006\\
Ours-C-100 & 0.753 $\pm$ 0.013 & 0.364 $\pm$ 0.035 & 0.003 & -0.017\\% & 0.982 $\pm$ 0.004 & 0.951 $\pm$ 0.007\\
% \textbf{U-Net} (12F)   & DSC-F & DSC-T & GAIN-$\mu$ &  GAIN-$\sigma$ \\ \hline %& AUC-F & AUC-T          \\ \hline
% F50-T50   & 0.777 $\pm$ 0.007 & 0.562 $\pm$ 0.022 & 0.0 & 0.0\\ % & 0.984 $\pm$ 0.004 & 0.963 $\pm$ 0.009\\
% F10-T90   & 0.745 $\pm$ 0.008 & 0.565 $\pm$ 0.016 & -0.029 & 0.005\\% & 0.986 $\pm$ 0.003 & 0.965 $\pm$ 0.009\\
% F90-T10   & 0.782 $\pm$ 0.010 & 0.497 $\pm$ 0.025 & -0.060 & -0.006\\% & 0.983 $\pm$ 0.004 & 0.921 $\pm$ 0.023\\\hline
% Simple-G   & 0.731 $\pm$ 0.035 & 0.501 $\pm$ 0.109 & -0.107 & -0.115\\% & 0.978 $\pm$ 0.006 & 0.926 $\pm$ 0.052\\
% Simple-C   & 0.756 $\pm$ 0.022 & 0.556 $\pm$ 0.034 & -0.027 & -0.027\\% & 0.984 $\pm$ 0.005 & 0.955 $\pm$ 0.020\\\hline
% Ours-G-25  & 0.777 $\pm$ 0.007 & 0.560 $\pm$ 0.014 & -0.002 & \textbf{0.008 }\\% & 0.984 $\pm$ 0.004 & 0.962 $\pm$ 0.010\\
% Ours-G-100 & 0.777 $\pm$ 0.007 & 0.559 $\pm$ 0.014 & -0.003 & \textbf{0.008}\\% & 0.985 $\pm$ 0.006 & 0.965 $\pm$ 0.007\\
% Ours-C-25  & 0.774 $\pm$ 0.009 & 0.564 $\pm$ 0.014 & -0.001 & 0.006\\% & 0.985 $\pm$ 0.003 & 0.967 $\pm$ 0.008\\
% Ours-C-100 & 0.777 $\pm$ 0.008 & 0.564 $\pm$ 0.018 & \textbf{0.002} & 0.003\\% & 0.985 $\pm$ 0.003 & 0.968 $\pm$ 0.008\\ \hline\hline  
\hline\hline\textbf{U-Net} (8F)   & DSC-F & DSC-T & GAIN-$\mu$ &  GAIN-$\sigma$\\\hline %  & AUC-F & AUC-T            \\ \hline
F50-T50   & 0.767 $\pm$ 0.013 & 0.556 $\pm$ 0.025 & 0.0 & 0.0\\% & 0.982 $\pm$ 0.007 & 0.966 $\pm$ 0.007 \\
F10-T90   & 0.745 $\pm$ 0.014 & 0.574 $\pm$ 0.017 & -0.004 & 0.007\\% & 0.984 $\pm$ 0.008 & 0.966 $\pm$ 0.009\\
F90-T10   & 0.775 $\pm$ 0.011 & 0.499 $\pm$ 0.028 & -0.049 & -0.001\\% & 0.980 $\pm$ 0.005 & 0.925 $\pm$ 0.020 \\\hline
Simple-G   & 0.745 $\pm$ 0.030 & 0.498 $\pm$ 0.129 & -0.080 & -0.121\\% & 0.981 $\pm$ 0.008 & 0.937 $\pm$ 0.046 \\
Simple-C   & 0.750 $\pm$ 0.015 & 0.555 $\pm$ 0.025 & -0.018 & -0.002\\% & 0.982 $\pm$ 0.006 & 0.964 $\pm$ 0.009 \\\hline
Ours-G-25  & 0.769 $\pm$ 0.014 & 0.555 $\pm$ 0.022 & 0.001 & 0.002\\% & 0.980 $\pm$ 0.008 & 0.960 $\pm$ 0.013 \\
Ours-G-100 & 0.769 $\pm$ 0.012 & 0.545 $\pm$ 0.022 & -0.009 & 0.004\\% & 0.981 $\pm$ 0.007 & 0.960 $\pm$ 0.011 \\
Ours-C-25  & 0.768 $\pm$ 0.014 & 0.566 $\pm$ 0.012 & \textbf{0.011} & \textbf{0.012}\\% & 0.985 $\pm$ 0.003 & 0.966 $\pm$ 0.010 \\
Ours-C-100 & 0.771 $\pm$ 0.009 & 0.561 $\pm$ 0.020 & 0.009 & 0.009\\% & 0.980 $\pm$ 0.008 & 0.963 $\pm$ 0.009 \\ \hline\hline  
% \textbf{LW} (8F)  & DSC-F & DSC-T & GAIN-$\mu$ &  GAIN-$\sigma$\\\hline % & AUC-F & AUC-T            \\ \hline
% F50-T50   & 0.753 $\pm$ 0.008 & 0.361 $\pm$ 0.023 & 0.0 & 0.0\\% & 0.982 $\pm$ 0.004 & 0.952 $\pm$ 0.007\\
% % F10-T90   &           &        &           &        &           &\\
% % F90-T10   &           &        &           &        &           &\\\hline
% % Simple-G   &           &        &           &        &            &\\
% % Simple-C   &           &        &           &        &           &\\\hline
% % Ours-G-25  &           &        &           &        &           &\\
% Ours-G-100 & 0.756 $\pm$ 0.010 & 0.368 $\pm$ 0.030 & \textbf{0.010} & -0.009\\% & 0.984 $\pm$ 0.004 & 0.956 $\pm$ 0.005\\
% Ours-C-25  & 0.752 $\pm$ 0.007 & 0.355 $\pm$ 0.021 & -0.007 & \textbf{0.003}\\% & 0.983 $\pm$ 0.006 & 0.953 $\pm$ 0.006\\\hline\hline
% % Ours-C-25  &           &        &           &        &           &\\
% % Ours-C-100 &           &        &           &        &           &\\ \hline\hline
\specialrule{.1em}{.1em}{0.1em} 
\end{tabular}
}
\end{table}

\vspace{0.5em}
\noindent\textbf{Metrics.} We evaluated the methods with the Dice Similarity Coefficient (DSC = 2TP / (2TP + FP + FN)).%, commonly used for segmentation evaluation (TP=True Pos., FP=False Pos., FN=False Neg.).
% We report the mean and s.d. of these across 5 repeated runs, each with different weight initializations.
\subsection{Results and Analyses}
% \begin{figure}
% %   \begin{center}
% \centering
%     \includegraphics[width=.6\columnwidth]{lambdas_compact.png}
% %   \end{center}
% \vspace{-10pt}
%   \caption{\label{fig:lambdas}\footnotesize Visualization of how Ours-C-25 modifies $\lambda_\theta$ over time (U-Net 8F). Solid line shows mean. Band shows s.d. Early spikes favor FLAIR, followed by T1-MR, then gradual return to even favor.}
%   \vspace{-5pt}
% \end{figure}
Table \ref{table:full} shows the results of all methods under various setups. We emphasize that our approach only modifies the baseline by allowing a dynamic weighting of the two domains. Therefore, our approach is intended to be a simple \textit{add-on} to the weighted loss approach and we do not expect staggering performance jumps in all cases. Instead, we hypothesize our method will improve upon the baseline in \textbf{low resource situations} (e.g., using less of the more informative FLAIR samples and the much smaller \textbf{LW} network). To this end, we show \textbf{U-Net} (8F) to demonstrate improvement when the number of FLAIR samples is down-sampled but the network is still large. We also show \textbf{LW} (8F) and \textbf{LW} (12F) to show two cases where the network is very under-parameterized. 

In all of these cases, our proposed approach demonstrates improvement over the compared baselines. Unlike ours, fixed weight setups (F10-T90 and F90-T10) are able to improve DSC on a single domain, but inevitably sacrifice performance on the others (i.e., giving worse overall performance). Fig.~\ref{fig:lambdas} emphasizes the importance of an adaptive weighting, showing how $\lambda_t$ is modified throughout training. But, \textit{naive} adaptive weighting may still fail. Poor performances of simple heuristics (Simple-G and Simple-C) show that $\lambda_t$ needs to be \textit{learned} as in our methods. Besides increased performance in DSC gain, our method also reduces the variability of the results across runs. In low data regimes, standard-deviation in performance during cross-validation can be very large -- our reduction in this measure indicates robustness to difficulty of the testing data and quality of the training data. 

% setup 1.2 page, results 1.5 page
\vspace{-3pt}
\section{Conclusion}
% \vspace{-8pt}
We proposed a model-agnostic solution to the problem of MDL. The solution is an extension of a simple weighted loss which uses meta-learning with inner-loop MAP Estimation to dynamically learn the weights of our loss function. On a WMH segmentation problem, we show that our proposed method improves both performance and consistency in low resource scenarios. The approach is widely applicable for MDL, making \textit{no} assumptions on the underlying model.     % 0.3 page
% \clearpage
% \noindent\textbf{Acknowledgments.} This work was supported by the NIH/NIA grants:
% R01 AG063752, RF1 AG025516, P01 AG025204, and SCI Undergraduate Research 
% Scholars Award. We report no conflicts of interests.

\section{Acknowledgments}
% \vspace{-12pt}
This work was supported by the NIH/NIA 
(R01 AG063752, RF1 AG025516, P01 AG025204, K23 MH118070), and SCI UR 
Scholars Award. We report no conflicts of interests.
% \vspace{-8pt}

% \vspace{-10pt}
\section{Compliance with Ethical Standards}
% \vspace{-8pt}
The study was performed in line with the principles of the Declaration of Helsinki. Approval was granted by the Ethics Committee of the University of Pittsburgh.
% supplement                % 2 page
%%%%%%%%%%%%%%%%%%%%%%%%%%%%%

% \newpage
\bibliographystyle{IEEEbib}
\bibliography{main}
\end{document}

\title{Multi-Domain Learning by Meta-Learning: Taking Optimal Steps in Multi-Domain Loss Landscapes by Inner-Loop Learning (Supplementary Material)}
\titlerunning{Multi-Domain Learning by Meta-Learning}
% \author{ID 1792}
% \institute{Anonymous}
%
\maketitle              % typeset the header of the contribution
\section{Loss Function}
For both FLAIR and T1-MR we minimize the loss-function below
\begin{equation}\small
    \frac{1}{n} \left ( \sum_{i=1}^n  \mathbf{y}_i\log(\mathbf{p}_i) + (1 - \mathbf{y}_i)\log(1 - \mathbf{p}_i) \right ) - \frac{\mathbf{y}^\mathrm{T}\mathbf{p}}{\sum_i \mathbf{y}_i + \sum_i \mathbf{p}_i}
\end{equation}
where $\mathbf{y}$ is the vector of per-pixel ground-truths across a mini-batch and $\mathbf{p}$ is the model-given probability score per-pixel across a mini-batch. The first term corresponds to the usual cross-entropy, while the second term corresponds to the differentiable Dice Score loss \cite{DBLP:journals/corr/abs-1802-10508}, specifically, a variant of the original one \cite{MilletariNA16}.

\section{Additional Results}
\begin{table}[!h]
\centering\scriptsize\renewcommand{\arraystretch}{1.1}\setlength{\tabcolsep}{3pt}
    \caption{Means and standard deviations (s.d.) of metrics using U-Net and training on a single domain (FLAIR or T1-MR). Training setup is identical to description in the main text. Decreased DSC-F and DSC-T (even when the training domain matches the testing domain) indicates knowledge transfer occurs during the joint-training schemes described in the main text.}
    \label{tab:my_label}
    \begin{tabular}{l|cc||cc}
Domain & DSC-F & DSC-T & AUC-F & AUC-T            \\ \hline\hline
    FLAIR & 0.7715 $\pm$ 0.0116 & 0.0005 $\pm$ 0.0006 & 0.9829 $\pm$ 0.0045 & 0.4524 $\pm$ 0.0930 \\\hline
    T1-MR & 0.0007 $\pm$ 0.0013 & 0.5310 $\pm$ 0.0277 & 0.0532 $\pm$ 0.0285 & 0.9383 $\pm$ 0.0200\\\hline
    \end{tabular}
\end{table}
\begin{table}[!t]
\centering\scriptsize\renewcommand{\arraystretch}{1.1}\setlength{\tabcolsep}{3pt}
    \caption{Means and standard deviations (s.d.) of metrics across all setups using LW and 8 FLAIR subjects. -F and -T indicate the metrics are computed over FLAIR and T1-MR samples respectively. GAIN-$\mu$ is the total (summed) increase in DSC over the baseline F50-T50.  GAIN-$\sigma$ is the total decrease in s.d. of DSC from F50-T50. While LW performs poorly due to its heavy under-parameterization, results generally align with our discussion in the main-text.} %Again, this is particularly true in the reduced data-availability case (8F).}
    \label{tab:my_label}
    \begin{tabular}{l|cc||cc||cc}
\textbf{LW} (12F)  & DSC-F & DSC-T & GAIN-$\mu$ &  GAIN-$\sigma$ & AUC-F & AUC-T            \\ \hline
F50-T50   & 0.757 $\pm$ 0.011 & 0.360 $\pm$ 0.031 & 0.0 & 0.0 & 0.983 $\pm$ 0.004 & 0.952 $\pm$ 0.007\\
F10-T90   & 0.729 $\pm$ 0.006 & 0.404 $\pm$ 0.026 & 0.016 & 0.010 & 0.977 $\pm$ 0.006 & 0.954 $\pm$ 0.006\\
F90-T10   & 0.766 $\pm$ 0.008 & 0.278 $\pm$ 0.033 & -0.073 & 0.001 & 0.986 $\pm$ 0.004 & 0.946 $\pm$ 0.007\\\hline
Simple-G   & 0.740 $\pm$ 0.023 & 0.152 $\pm$ 0.072 & -0.225 & -0.053 & 0.977 $\pm$ 0.011 & 0.880 $\pm$ 0.052\\
Simple-C   & 0.714 $\pm$ 0.062 & 0.325 $\pm$ 0.050 & -0.078 & -0.070 & 0.976 $\pm$ 0.010 & 0.941 $\pm$ 0.020\\\hline
Ours-G-25  & 0.758 $\pm$ 0.009 & 0.366 $\pm$ 0.029 & 0.007 & 0.004 & 0.983 $\pm$ 0.003 & 0.952 $\pm$ 0.003\\
Ours-G-100 & 0.759 $\pm$ 0.010 & 0.375 $\pm$ 0.028 & \textbf{0.017} & 0.004 & 0.983 $\pm$ 0.003 & 0.952 $\pm$ 0.006\\
Ours-C-25  & 0.758 $\pm$ 0.007 & 0.356 $\pm$ 0.025 & -0.003 & 0.010 & 0.983 $\pm$ 0.003 & 0.953 $\pm$ 0.005\\
Ours-C-100 & 0.755 $\pm$ 0.008 & 0.351 $\pm$ 0.018 & -0.011 & \textbf{0.016} & 0.982 $\pm$ 0.004 & 0.951 $\pm$ 0.008\\ \hline\hline  
\textbf{LW} (8F)  & DSC-F & DSC-T & GAIN-$\mu$ &  GAIN-$\sigma$ & AUC-F & AUC-T            \\ \hline\hline
F50-T50   & 0.753 $\pm$ 0.008 & 0.361 $\pm$ 0.023 & 0.0 & 0.0 & 0.982 $\pm$ 0.004 & 0.952 $\pm$ 0.007\\
F10-T90   & 0.725 $\pm$ 0.008 & 0.393 $\pm$ 0.026 & 0.004 & -0.003 & 0.975 $\pm$ 0.006 & 0.95 $\pm$ 0.009\\
F90-T10   & 0.766 $\pm$ 0.013 & 0.291 $\pm$ 0.030 & -0.057 & -0.012 & 0.986 $\pm$ 0.004 & 0.948 $\pm$ 0.007\\\hline
Simple-G   & 0.738 $\pm$ 0.020 & 0.152 $\pm$ 0.055 & -0.224 & -0.044 & 0.978 $\pm$ 0.009 & 0.881 $\pm$ 0.067\\
Simple-C   & 0.716 $\pm$ 0.063 & 0.311 $\pm$ 0.081 & -0.087 & -0.113 & 0.975 $\pm$ 0.014 & 0.935 $\pm$ 0.033\\\hline
Ours-G-25  & 0.755 $\pm$ 0.007 & 0.361 $\pm$ 0.023 & 0.002 & 0.001 & 0.982 $\pm$ 0.004 & 0.952 $\pm$ 0.007\\
Ours-G-100 & 0.756 $\pm$ 0.010 & 0.368 $\pm$ 0.030 & \textbf{0.010} & -0.009 & 0.984 $\pm$ 0.004 & 0.956 $\pm$ 0.005\\
Ours-C-25  & 0.752 $\pm$ 0.007 & 0.355 $\pm$ 0.021 & -0.007 & \textbf{0.003} & 0.983 $\pm$ 0.006 & 0.953 $\pm$ 0.006\\
Ours-C-100 & 0.753 $\pm$ 0.013 & 0.364 $\pm$ 0.035 & 0.003 & -0.017 & 0.982 $\pm$ 0.004 & 0.951 $\pm$ 0.007\\ \hline\hline
    \end{tabular}
\end{table}
% \begin{tabular}{l|cc||cc||cc} \hline
% % \textbf{LW} (8F)  & DSC-F & DSC-T & GAIN-$\mu$ &  GAIN-$\sigma$ & AUC-F & AUC-T            \\ \hline
% % F50-T50   &           &        &           &        &           &\\
% % F10-T90   &           &        &           &        &           &\\
% % F90-T10   &           &        &           &        &           &\\\hline
% % Simple-G   &           &        &           &        &            &\\
% % Simple-C   &           &        &           &        &           &\\\hline
% % Ours-G-25  &           &        &           &        &           &\\
% % Ours-G-100 &           &        &           &        &           &\\
% % Ours-C-25  &           &        &           &        &           &\\
% % Ours-C-100 &           &        &           &        &           &\\ \hline\hline    
% \begin{tabular}
%%%%%%%%%%%%%%%%%%%%%%%%%%%%%

\newpage
\bibliographystyle{splncs04}
\bibliography{miccai2020}